\documentclass{article}

\usepackage[final,nonatbib]{neurips_2022}

\usepackage[utf8]{inputenc} %
\usepackage[T1]{fontenc}    %
\usepackage{hyperref}       %
\usepackage{url}            %
\usepackage{booktabs}       %
\usepackage{amsfonts}       %
\usepackage{nicefrac}       %
\usepackage{microtype}      %
\usepackage[pdftex]{graphicx}
\usepackage{amsmath}
\usepackage{wrapfig}
\usepackage{multirow}
\usepackage{caption}
\usepackage{subcaption}
\usepackage{enumitem}
\usepackage[dvipsnames]{xcolor}

\newcommand{\modelname}{ViL3DRel}

\def\onedot{. }

\def\etal{\emph{et al}\onedot}

\title{Language Conditioned Spatial Relation Reasoning\\ for 3D Object Grounding}

\author{Shizhe Chen$^{1}$, Pierre-Louis Guhur$^{1}$, Makarand Tapaswi$^{2}$, Cordelia Schmid$^{1}$, Ivan Laptev$^{1}$\\[3pt]
$^{1}$Inria, \'Ecole normale sup\'erieure, CNRS, PSL Research University, $^{2}$IIIT Hyderabad\\ [3pt]
{\small \url{https://cshizhe.github.io/projects/vil3dref.html}}}

\begin{document}

\maketitle

\begin{abstract}
Localizing objects in 3D scenes based on natural language requires understanding and reasoning about spatial relations. 
In particular, it is often crucial to distinguish similar objects referred by the text, such as "the left most chair" and "a chair next to the window".
In this work we propose a language-conditioned transformer model for grounding 3D objects and their spatial relations.
To this end, we design a spatial self-attention layer that accounts for relative distances and orientations between objects in input 3D point clouds.
Training such a layer with visual and language inputs enables to disambiguate spatial relations and to localize objects referred by the text.
To facilitate the cross-modal learning of relations, we further propose a teacher-student approach where the teacher model is first trained using ground-truth object labels, and then helps to train a student model using point cloud inputs. 
We perform ablation studies showing advantages of our approach.
We also demonstrate our model to significantly outperform the state of the art on the challenging Nr3D, Sr3D and ScanRefer 3D object grounding datasets.
\end{abstract}

\section{Introduction}

To carry out human instructions in the real world, robots should  understand natural language and be able to ground mentioned objects in 3D environments. 
Following this objective, recent research is shifting from object grounding in 2D images~\cite{yu2016modeling,mao2016generation,liu2019clevr,chen2020cops,chen2019touchdown,qi2020reverie} to the 3D object grounding task~\cite{chen2020scanrefer,achlioptas2020referit3d}, where objects referred by a sentence should be localized in a 3D point cloud.

\begin{figure}
     \centering
     \begin{subfigure}[b]{0.45\linewidth}
         \centering
         \includegraphics[width=\textwidth]{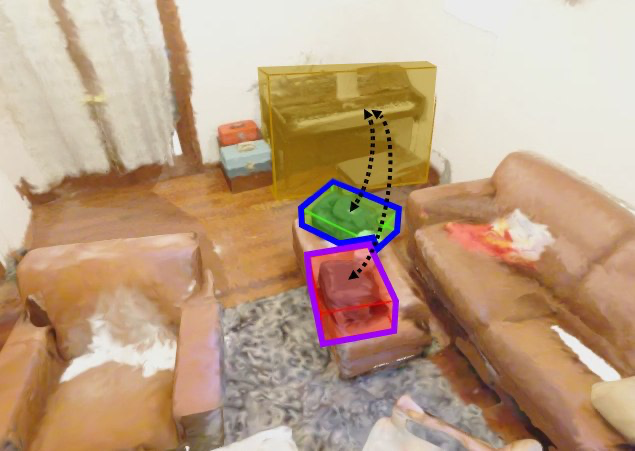}
         \caption{The \textcolor{ForestGreen}{backpack} closest to the \textcolor{BurntOrange}{piano}.}
         \label{fig:intro_example_0}
     \end{subfigure}
     \hfill
     \begin{subfigure}[b]{0.4\linewidth}
         \centering
         \includegraphics[width=\linewidth]{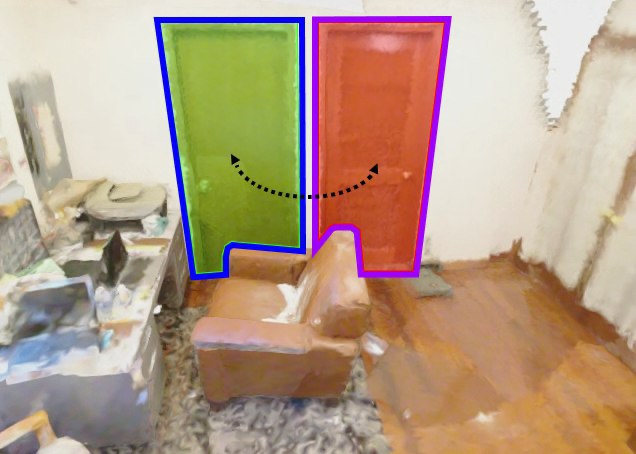}
         \caption{Of the two brown wooden doors, choose the \textcolor{ForestGreen}{door} on the left when facing them.}
         \label{fig:intro_example_1}
     \end{subfigure}
    \caption{Example sentences that refer to objects in  3D scenes. The green box denotes the ground-truth object, the blue box is the prediction from our model, and the purple one is from a baseline model without explicit spatial reasoning and knowledge distillation.}
    \label{fig:intro}
    \vspace{-1em}
\end{figure}

Language expressions often refer to objects by their relative spatial locations in 3D scenes.
Figure~\ref{fig:intro} illustrates example scenes and corresponding sentences where object grounding requires disambiguation between objects of the same class.  
For instance, \emph{``the backpack closest to the piano''} requires to compare relative distances among objects, while \emph{``choose the door on the left when facing them''} requires to infer the correct viewpoint and understand relative directions.
Such complexity and diversity of the spatial language makes 3D object grounding highly challenging.

Given the critical role of the spatial language, many existing methods attempt to model  spatial relations for 3D object grounding.
Early work \cite{he2021transrefer3d,huang2021text,yuan2021instancerefer,feng2021free} explicitly build 3D visual graphs based on distances between objects and apply graph neural networks to learn relationships.
However, since only nearest neighbors are considered in the graph, it remains difficult to infer relationships between distant objects such as \emph{``farthest''}.
More recently, transformer architectures \cite{he2021transrefer3d,zhao20213dvg,yang2021sat,roh2022languagerefer,huang2022multi} have been used, as they  have the potential to learn relations between pairs of objects with a multi-head self-attention mechanism \cite{vaswani2017attention}.
Enabling transformers to better understand 3D spatial relations expressed by natural language, however, remains an open research problem.

In addition, 3D object grounding suffers more from the scarcity of training data compared to its 2D counterpart.
The success of 2D object grounding models \cite{kamath2021mdetr,subramanian2022reclip} can be largely attributed to large-scale datasets with image-text pairs \cite{yu2016modeling,mao2016generation,krishna2017visual,radford2021learning}, 
whereas the limited amount of 3D scene-language pairs increases the  difficulty of 3D object reasoning. %
To address this challenge, Yang \etal \cite{yang2021sat} propose to use 2D-3D alignments to assist the training of 3D models, but require additional high-quality 2D images and camera parameters which are not always available.

In this work, we propose a {\bf Vi}sion-and-{\bf L}anguage {\bf 3D} {\bf Rel}ation reasoning model (\modelname) to tackle the above issues in 3D object grounding.
We design a new spatial self-attention layer for the transformer architecture to enhance 3D spatial understanding.
This layer encodes relative distances and orientations for all pairs of objects, and explicitly learns spatial attention to capture spatial relations referred by the the language expressions.
We propose a sigmoid softmax function to effectively fuse the spatial attention with standard self-attention.
Rotation augmentation is used to further improve view invariant reasoning.
To alleviate the negative effect of noisy object features, we further propose a teacher-student training approach. The teacher and student share the same model architecture, but use different inputs. The teacher uses ground-truth object labels, which enables to better learn the relation reasoning. 
The learned relation knowledge is distilled to the student taking point cloud features as input.
Our code and models are available on the project webpage~\cite{projectpage}.

To summarize, our contributions are three-fold:
\parskip=0.1em
\begin{itemize}[itemsep=0.1em,parsep=0em,topsep=0em,partopsep=0em]
    \item We propose a \modelname~model for the 3D object grounding task. It uses a new spatial self-attention to explicitly encode pairwise 3D spatial relations in the transformer layer for better language conditioned spatial understanding.
    \item We employ a teacher-student training strategy, which facilitates the cross-modal learning of relations. The student with point cloud inputs benefits from the relation reasoning model of the teacher 
    trained with ground-truth object label input.
    \item We evaluate our \modelname~approach on the challenging Nr3D, Sr3D \cite{he2021transrefer3d} and ScanRefer \cite{chen2020scanrefer} benchmarks. Our model significantly outperforms state-of-the-art approaches, with 9.3, 8.3 absolute gains on Nr3D and Sr3D given ground-truth object proposals, and 4.47 points on ScanRefer using the same detected object proposals compared to the previous work \cite{huang2022multi} . 
\end{itemize}
\section{Related Work}

\noindent\textbf{2D and 3D Object Grounding.}
The object grounding task aims to localize objects in 2D images or 3D point clouds given a sentence.
Existing approaches can be categorized into two groups, namely  one- and two-stage frameworks.
The one-stage methods densely fuse the text features with patch- or point-level visual representations to directly regress the bounding box \cite{kamath2021mdetr,luo20223dsps}. These approaches are flexible to detect various objects given the input sentence.
The two-stage methods adopt a \emph{detection-then-matching} pipeline \cite{yu2016modeling,chen2020scanrefer,achlioptas2020referit3d,yu2018mattnet}, where the detection stage generates object proposals, and then the matching stage selects the best proposal according to the sentence.
The decoupled object perception and cross-modal matching make the two-stage methods easier to analyze %
\cite{roh2022languagerefer}. 

The benchmark datasets for 3D object grounding are ReferIt3D (Nr3D and Sr3D)~\cite{achlioptas2020referit3d} and ScanRefer~\cite{chen2020scanrefer}, which are all built upon scenes and object annotations in the ScanNet dataset \cite{dai2017scannet}.
Graph-based approaches \cite{velivckovic2018graph,wang2019dynamic} are widely adopted in early works to infer spatial relations.
The object graph is constructed by connecting each object with its top nearest neighbors \cite{achlioptas2020referit3d,huang2021text,yuan2021instancerefer} based on Euclidean distance.
Inspired by the success of transformers \cite{vaswani2017attention}, recent works \cite{he2021transrefer3d,yang2021sat,roh2022languagerefer,huang2022multi,luo20223dsps} have adopted transformers for 3D object grounding.
BEAUTY-DETR \cite{jain2021looking} and 3D-SPS \cite{luo20223dsps} are one-stage methods.
However, most works follow the two-stage framework with pre-detected object proposals.
LanguageRefer \cite{roh2022languagerefer} converts the cross-modal task into a language modeling problem with predicted object labels.
SAT \cite{yang2021sat} adopts a multimodal transformer and transfers 2D semantics to assist the training of the 3D model.
Multi-view transformer \cite{huang2022multi} aggregates object representations from multiple views to improve view robustness.
Perhaps most similar to our work, 3DVG-Transformer \cite{zhao20213dvg} proposes a coordinate-guided attention module to encode spatial distances among objects. In contrast, we introduce a spatial self-attention module, which explicitly encodes both relative distances and relative orientations among objects. It also conditions spatial relations on the language and presents a more effective strategy for attention fusion.

\noindent\textbf{Transformers in Vision-and-Language.}
Transformer-based architectures have led to significant improvements in various vision-and-language tasks such as text-video retrieval~\cite{sun2019videobert}, visual grounding~\cite{kamath2021mdetr,yang2022tubedetr}, image captioning~\cite{cornia2020meshed}, vision question-answering \cite{zhou2020unified} and vision-and-language navigation~\cite{chen2021history,chen2022think}.
Most of the methods project textual and visual inputs into a sequence of tokens, and use multimodal transformers to learn cross-modal semantic relationships.
While relative positional encoding (RPE)~\cite{wu2021rethinking} has been explored mostly separately in vision and language transformers, we here integrate RPE with a cross-modal transformer and show its benefits for resolving spatial object relations referred by text.  

\noindent\textbf{Knowledge Distillation.}
Knowledge distillation \cite{hinton2015distilling} is typically used to compress a large network (teacher) into a compact model (student).
The common objective is to let the student model mimic the soft logits of the teacher model.
Beyer \etal \cite{beyer2021knowledge} show that a consistent and patient teacher is essential in knowledge distillation.
Jiao \etal \cite{jiao2020tinybert} further demonstrate that the intermediate representations learned by the teacher are beneficial. %
In contrast to distilling knowledge from a heavy model to a light one, our teacher network first learns cross-modal object relations using ground-truth object labels. We then transfer this knowledge to the target student network that uses noisy inputs.

\section{Method}

Given a sentence $S$, the goal of 3D object grounding is to detect an object referred in $S$ by locating its 3D bounding box $B_T \in \mathbb{R}^6$ in a 3D point clouds $P_{scene}$.
We assume $P_{scene} \in \mathbb{R}^{K \times 6}$ contains $K$ points each represented by XYZ coordinates and RGB values.
We follow the \emph{detection-then-matching} framework \cite{chen2020scanrefer,achlioptas2020referit3d,zhao20213dvg,yang2021sat,huang2022multi} for 3D object grounding and assume to be given a list of object proposals $(O_1, \cdots, O_N)$ obtained via automatic 3D instance segmentation or ground-truth annotations (depending on the evaluation setup). 
Each object $O_i$ is represented by a subset of $K_i$ points $P_i \subset P_{scene}, P_i \in \mathbb{R}^{K_i \times 6}$.
Our work is focused on interpreting spatial relations and selecting the target object $O_T, T \in [1, N]$ among $N$ object proposals.
We first present an overview of our \modelname~model in Section~\ref{sec:method_3dog_model}.
We then introduce our language-conditioned spatial self-attention module and the teacher-student training approach in Sections~\ref{sec:method_spatial} and ~\ref{sec:method_training} respectively.

\subsection{Architecture Overview}
\label{sec:method_3dog_model}
\begin{figure}
    \centering
    \includegraphics[width=\linewidth]{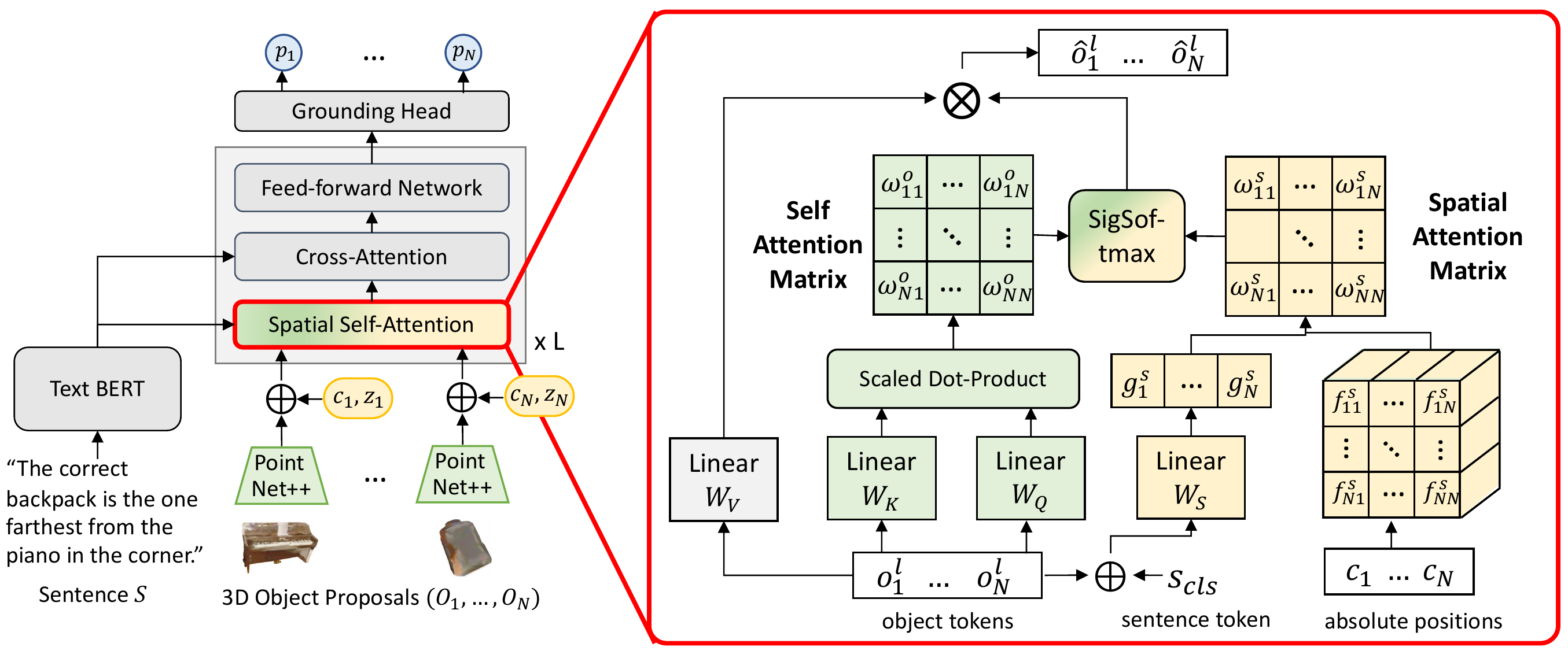}
    \caption{Left: overview of the \modelname~model; Right: language-conditioned spatial self-attention.}
    \label{fig:3dog_model}
    \vspace{-1em}
\end{figure}

Figure~\ref{fig:3dog_model} shows an overview of our \modelname~model, consisting of four modules: text encoding, object encoding, multimodal fusion and a grounding head as described next.

\noindent\textbf{Text encoding.}  
Given the sentence $S$ with $M$ word tokens, we use a pre-trained BERT model \cite{devlin2019bert} to encode $S$ into a sequence of word features $(s_{cls}, s_1, \cdots, s_M), s_i \in \mathbb{R}^{d}$, where $s_{cls}$ is a special classification token and $d$ is the dimensionality of the feature.

\noindent\textbf{Object encoding.}
For each object $O_i$, we first normalize the coordinates of its point cloud $P_i$ into a unit ball, and then use PointNet++ \cite{qi2017pointnet++} to compute the object feature $o^0_i \in \mathbb{R}^d$.
We also obtain the object center $c_i = [c_x, c_y, c_z] \in \mathbb{R}^3$ and the object size $z_i = [z_x, z_y, z_z] \in \mathbb{R}^3$ from object points $P_i$ as the mean and the spatial extent of $P_i$ respectively.
We use a linear projection layer to obtain the absolute 3D location feature as $l_i = W_l [c_i; z_i] \in \mathbb{R}^d$.

\noindent\textbf{Multimodal fusion.}
A stack of transformer layers \cite{vaswani2017attention} are applied to fuse the text and object modality features.
Each transformer layer is composed of a spatial self-attention layer, a cross-attention layer and a feed-forward neural network (FFN).
Our new spatial self-attention layer aims to improve the understanding of spatial relations among objects referred by the sentence.
Assume $o^l_i$ is the input embedding for object proposal $O_i$ before the $i$-th layer, we first add it with its absolute 3D location feature $l_i$ to enhance the spatial information. The spatial self-attention then generates contextualized object representations $\hat{o}^l_i$.
We will describe the proposed spatial self-attention in details in Section~\ref{sec:method_spatial}.
The cross-attention layer takes $\hat{o}^l_i$ as queries and text features $(s_{cls}, s_1, \cdots, s_M)$ as keys and values to learn cross-modal relations.
The final FFN uses two fully connected layers to encode each output tokens after the attention layer.
For detailed explanation of Transformer attention and FFN layers please refer to~\cite{vaswani2017attention}.

\noindent\textbf{Object grounding.}
We use a two-layer feed-forward neural network as the object grounding head to predict the target object.
Given the output embedding $o^L_i$ from the last multimodal fusion layer, the grounding head generates a scalar score for each object proposal $O_i$ and applies softmax function to obtain the probability $p_i$.
The object proposal with the maximum probability is predicted as target.

\subsection{Spatial Self-Attention}
\label{sec:method_spatial}

The standard self-attention relies on token embeddings to learn relations among tokens.
However, the input embeddings of object proposals are mixed with semantic features and absolute 3D locations, making it difficult to accurately infer spatial relations among objects such as relative distances or orientations in Figure~\ref{fig:intro}.
As spatial relations among objects referred by language are important to distinguish object instances, we propose to inject a language conditioned spatial attention that explicitly captures pairwise spatial relations to complement the standard self-attention.
We first describe the standard self-attention and then introduce our proposed spatial self-attention as illustrated in the right part of Figure~\ref{fig:3dog_model}.

Given the feature matrix $X \in \mathbb{R}^{N \times d}$ for $N$ object proposals, the self-attention mechanism first computes the query, key and value embeddings from $X$ as $Q=XW_Q, K=XW_K, V=XW_V$ respectively, where $W_Q, W_K, W_V \in \mathbb{R}^{d \times d_h}$ are learnable parameters and $d_h$ is the dimensionality of the output embedding.
It then calculates attention weights given the query and key embeddings and aggregates the value embeddings as follows:
\begin{equation}
\label{eqn:self_attn}
	\Omega^o = \text{softmax}\left(\frac{Q K^T}{\sqrt{d_h}}\right); \quad \text{SelfAttn}(Q, K, V) = \Omega^o V,
\end{equation}
where $\Omega^o$ is an $N \times N$ attention matrix, whose elements $\omega^o_{ij}$ is the attention weight between the $i$-th and $j$-th object proposal.
In order to capture more diverse relations, multi-head self-attention is used where each head computes an independent $\text{SelfAttn}(Q, K, V)$ and the outputs from all heads are concatenated.

To model spatial relations among objects, we propose to use explicit pairwise spatial features $f^s_{ij} \in \mathbb{R}^5, i,j \in [1, N]$.
For each pair of objects $(O_i, O_j)$, we compute their Euclidean distance $d_{ij}=||c_i-c_j||_2$
as well as horizontal and vertical angles $\theta_h, \theta_v$ of the line connecting object centers $c_i$ and $c_j$.
The computation details of the angles are provided in the Section~\ref{sec:supp_model} of supplementary material.
We then define the pairwise spatial feature $f^s_{ij}$  as:
\begin{equation}
\label{eqn:pairwise_spatial_features}
	f^s_{ij} = [d_{ij}, \sin(\theta_h), \cos(\theta_h), \sin(\theta_v), \cos(\theta_v)].
\end{equation}
We automatically generate a language conditioned weight $g^s_{i}$ to select relevant spatial relations for each object proposal $O_i$ as we mainly care about spatial relations described in the text for an object:
\begin{equation}
	g^s_{i} = W_S^T (s_{cls} + o^l_i),
\end{equation}
where $W_S \in \mathbb{R}^{d \times 5}$ is a learnable parameter and we omit the bias term for simplicity.
We then define the spatial relevance for $(O_i, O_j)$ as:
\begin{equation}
	\omega^{s}_{ij} = g^s_{i} \cdot f^s_{ij}.
\end{equation}
We modulate the self-attention in Eq~(\ref{eqn:self_attn}) with the above language conditioned spatial relevancy using the sigmoid softmax (sigsoftmax) fusion function:
\begin{equation}
\label{eqn:spatial_self_attn}
	\omega_{ij} = \frac{\sigma(\omega^s_{ij})~\text{exp}(w^o_{ij})}{\sum_{l=1}^{N} \sigma(\omega^s_{il})~\text{exp}(w^o_{il})},
\end{equation}
where $\sigma(\cdot)$ is the sigmoid function.
In this way, we compute the new self-attention matrix $\Omega=[\omega_{ij}]_{N \times N}$ which explicitly considers 3D relative spatial locations, absolute spatial locations and object appearances in relation reasoning.
We also use the multi-head mechanism to support different types of spatial relations and compute multiple spatial attentions to fuse with the standard self-attention. We concatenate the outputs from all heads in the spatial self-attention layer.

\subsection{Teacher-Student Training}
\label{sec:method_training}

\begin{figure}%
	\centering
	\includegraphics[width=0.9\linewidth]{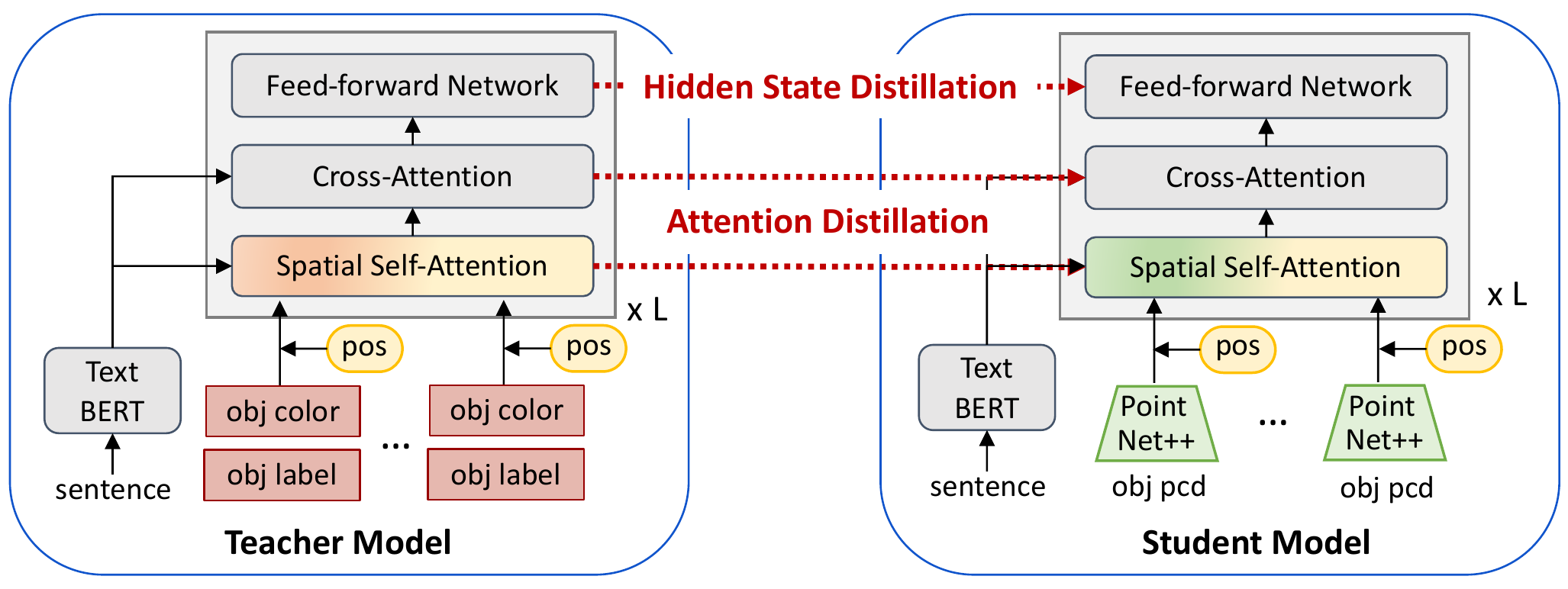}
	\caption{Teacher student learning with hidden state and attention distillation.}
	\label{fig:teacher_student_train}
	\vspace{-1em}
\end{figure}

Incorrect estimation of object classes from $P_i$ can deteriorate the cross-modal alignment and spatial relation reasoning.
To improve object grounding in point clouds previous work~\cite{yang2021sat} relies on additional supervision in 2D images. 
Here, instead, we propose a new teacher-student training approach using no additional training data.
The teacher and student models share the same transformer architectures except for the object encoding module.
The teacher uses ground-truth object semantic features, while the student uses 3D point clouds. 
Since there is less cross-modal gap between the teacher's inputs and the sentence, the teacher model can excel at learning language relevant spatial relations among objects and aligning objects with the sentence.
Such knowledge can then be distilled to the student model to improve its training.
Figure~\ref{fig:teacher_student_train} illustrates the teacher-student training pipeline. 
We describe the inputs of the teacher and the knowledge distillation objectives in the following.

\noindent\textbf{Teacher's inputs.}
We use ground-truth class labels and dominant colors of the object as the object representations for the teacher model, since the color is the most widely used attributes in the sentence.
Specifically, we encode the ground-truth object class label using pre-trained Glove word vectors \cite{pennington2014glove}.
To obtain the dominant colors of the object, we fit a Gaussian Mixture model on the RGB values of all points in the object, where the mixture component is set to 3.
We linearly project the mean value in each component into a color embedding and use the mixture weights of the components to average all the color embeddings.
The final object representation is the sum of the class label embedding and the averaged color embedding.

\noindent\textbf{Knowledge distillation objectives.}
We transfer the attention weights and hidden states in the multimodal fusion module of the teacher model to the student model.
The self-attentions learned by the teacher capture spatial relations among objects, and the cross-attentions measure cross-modal matching, which are both essential for 3D object grounding but hard to learn given noisy object features.
We facilitate student learning by forcing it to mimic all the attention matrices $\Omega$ of the teacher using the following objective function:
\begin{equation}
\label{eqn:attn_distill_loss}
	\mathcal{L}_{attn} = \frac{1}{LH} \sum_{l=1}^{L} \sum_{h=1}^{H} ~\text{MSE}(\Omega_{lh}^S - \Omega_{lh}^T),
\end{equation}
where $\Omega_{\cdot}^S$ and $\Omega_{\cdot}^T$ are attention weights in student and teacher model respectively, $\text{MSE}(\cdot)$ is the mean square error function, $L$ is the number of multimodal fusion layers and $H$ is the number of heads in self-attention.
In addition to distilling the relation knowledge in attention weights, we also distill the output embeddings of the transformer layers:
\begin{equation}
\label{eqn:hidden_distill_loss}
	\mathcal{L}_{hidden} = \frac{1}{LN} \sum_{l=0}^{L} \sum_{i=1}^{N} ~\text{MSE}(o_{i}^{lS} - o_{i}^{lT}).
\end{equation}

\noindent\textbf{Training.}
We follow the previous works \cite{achlioptas2020referit3d,chen2020scanrefer} to use multiple auxiliary losses in training the transformer model, including a 3D object grounding loss $\mathcal{L}_{og}$, sentence classification loss $\mathcal{L}_{sent}$ and two object classification losses $\mathcal{L}_{obj}^{u}$ and $\mathcal{L}_{obj}^{m}$.
$\mathcal{L}_{sent}$ relies on $s_{cls}$ to predict the target object class from the sentence. $\mathcal{L}_{obj}^{u}$ and $\mathcal{L}_{obj}^{m}$ use unimodal object representation $o^0_i$ and multimodal fused object representation $o^L_i$ respectively to predict the classes for input object proposals.
Details are in Section~\ref{sec:supp_model} of supplementary material.
Therefore, the overall training objective is as follows:
\begin{equation}
\label{eqn:loss_func}
	\mathcal{L} =  \mathcal{L}_{og} + \mathcal{L}_{sent}  + \mathcal{L}_{obj}^{u} + \mathcal{L}_{obj}^{m} + \lambda_{a} \mathcal{L}_{attn} + \lambda_{h} \mathcal{L}_{hidden}
\end{equation}
where $\lambda_{a},  \lambda_{h}$ are two hyper-parameters to balance the losses.

\section{Experiments}

\subsection{Datasets}

\noindent\textbf{Nr3D dataset \cite{achlioptas2020referit3d}} contains 37,842 human-written sentences that refer to annotated objects in the 3D indoor scene dataset ScanNet~\cite{dai2017scannet}. The dataset includes 641 scenes with 511 (resp.~130) scenes for training (resp.~validation).
It covers 76 target object classes. The annotated sentences are designed to refer to objects with multiple same-class distractors in the scene.
The sentences are split into ``easy'' and ``hard'' subsets in evaluation, where the target object in ``easy'' subset only contains one same-class distractor in the scene while it contains multiple ones in the ``hard'' subset.
According to whether the sentence requires a specific viewpoint to ground the referred object, the dataset can also be partitioned into ``view depedent'' and ``view independent'' subsets.

\noindent\textbf{Sr3D dataset~\cite{achlioptas2020referit3d}} is constructed using templates to automatically generate sentences. The sentences only utilize spatial relations to distinguish objects of the same class.
It has 1,018 training scenes and 255 validation scenes from ScanNet and 83,570 sentences in total.
The dataset can be split in the same way as Nr3D during evaluation.

\noindent\textbf{ScanRefer dataset~\cite{chen2020scanrefer}} has 51,583 human-written sentences for 800 scenes in ScanNet. 
We follow the official split and use 36,665 and 9,508 samples for training and validation respectively.
According to whether the target object is a unique object class in the scene, the dataset can be divided into a ``unique'' and a ``multiple'' subset.

\subsection{Experimental Setting}

\noindent\textbf{Evaluation Metrics.}
We evaluate models under two evaluation settings.
One uses ground-truth object proposals, which is the default setting in the Nr3D and Sr3D datasets. The metric is the accuracy of selecting the target bounding box among the proposals.
The other setting does not provide ground-truth object proposals and requires the model to regress a 3D bounding box, which is the default setting for the ScanRefer dataset. The evaluation metrics are acc@0.25 and acc@0.5, which is the percentage of correctly predicted bounding boxes whose IoU is larger than 0.25 or 0.5 with the ground-truth. 

\noindent\textbf{Implementation details.}
For the model architecture, we set the dimension $d = 768$ and use 12 heads for all the transformer layers.
The text encoding module is a three-layer transformer initialized from BERT \cite{devlin2019bert}, and the multimodal fusion module contains four layers.
The object encoding module PointNet++~\cite{qi2017pointnet++} samples 1024 points for all the objects.
These architecture parameters are the same as in previous work~\cite{yang2021sat,huang2022multi} to ensure a fair comparison.
We first train the PointNet++ for object classification on ScanNet, which achieves 61.9\% accuracy on the validation set. Its weights remain fixed during the following training steps. 
Rotation augmentation is used to increase the viewpoint invariance.
The hyper-parameters in the loss function Eq~(\ref{eqn:loss_func}) are set to $\lambda_a = 1$ and $\lambda_h = 0.02$.
We train the model with a batch size of 128 and a learning rate of 0.0005 with warm-up of 5000 iterations and cosine decay scheduling. The AdamW algorithm \cite{loshchilov2018fixing} is used in the optimization.
We train for 50 epochs for the teacher model and 100 epochs for the student model on Nr3D and ScanRefer datasets.
The training epochs are reduced to half on the Sr3D dataset, as it is larger and easier to converge.
All models are trained on a single NVIDIA RTX A6000 GPU.

\subsection{Ablation Studies}
We carry out extensive experiments on the Nr3D dataset to demonstrate the effectiveness of our proposed spatial self-attention and teacher-student training.
The ablations on the ScanRefer dataset are provided in Section~\ref{sec:supp_results} of supplementary material.

\begin{table}
	\centering
	\caption{Grounding accuracy (\%) on the Nr3D dataset with ground-truth object labels. Dist stands for Distance; Ort for Orientation; MHA for multi-head spatial attention; RotAug for Rotation Augmentation; sigs for sigsoftmax in Eq~(\ref{eqn:spatial_self_attn}); and `-' means not applicable.}
	\label{tab:ablation_nr3d_spatial}
	\begin{tabular}{cccccccccc} \toprule
		\multirow{2}{*}{} & \multicolumn{4}{c}{Spatial Relation Reasoning} & \multirow{2}{*}{\begin{tabular}[c]{@{}c@{}}Rot\\ Aug\end{tabular}} & \multirow{2}{*}{Color} & \multirow{2}{*}{Overall} & \multirow{2}{*}{ViewDep} & \multirow{2}{*}{ViewIndep} \\ 
		& Dist & Ort & MHA & Fusion &  &  &  &  &  \\ \midrule
		R1 & - & - & - & - & $\times$ & $\times$ & 53.5 & 51.4 & 54.6 \\
		R2 & - & - & - & - & $\times$ & \checkmark & 55.1 & 53.8 & 55.8 \\
		R3 & - & - & - & - & \checkmark & \checkmark & 62.4 & 58.3 & 64.5 \\ \midrule
		R4 & \checkmark & $\times$ & \checkmark & sigs & \checkmark & \checkmark & 66.0 & 53.8 & 72.0 \\
		R5 & $\times$ & \checkmark & \checkmark & sigs & \checkmark & \checkmark & 71.3 & 68.5 & 72.6 \\
		R6 & \checkmark & \checkmark & $\times$ & sigs & \checkmark & \checkmark & 67.7 & 65.2 & 69.0 \\
		\midrule
		R7 & \checkmark & \checkmark & \checkmark & bias & \checkmark & \checkmark & 55.4 & 46.8 & 59.6 \\
		R8 & \checkmark & \checkmark & \checkmark & ctx & \checkmark & \checkmark & 56.4 & 50.8 & 59.1 \\ \midrule
		R9 & \checkmark & \checkmark & \checkmark & sigs & \checkmark & \checkmark & \textbf{74.4} & \textbf{71.3} & \textbf{75.9} \\ \bottomrule
	\end{tabular}
\end{table}

\subsubsection{Spatial Relation Reasoning}
We first evaluate the proposed spatial self-attention using ground-truth object labels, which decouples the object perception from spatial relation reasoning.
Table~\ref{tab:ablation_nr3d_spatial} presents results on the Nr3D dataset.

\noindent\textbf{Baselines.}
Our baseline R1 is the teacher model in the left part of Figure~\ref{fig:teacher_student_train} that excludes object color, rotation augmentation and the proposed spatial self-attention.
This baseline achieves similar performance to the state-of-the-art LanguageRefer \cite{roh2022languagerefer} (overall accuracy: 54.3\%) which also uses ground-truth object labels. 
Row R2 shows that adding color information helps object grounding as sentences often contain color attributes.
The rotation augmentation brings additional gains and improves accuracy from 55.1\% (R2) to 62.4\% (R3). It is more effective for view independent sentences (+8.7\%) than for view dependent sentences (+4.5\%) because rotation augmentation mainly improves view invariance.
We consider R3 as a strong baseline and use it to demonstrate improvements of our proposed model.

\noindent\textbf{Pairwise spatial features: distance vs. orientation.}
Rows R4-R5 in Table~\ref{tab:ablation_nr3d_spatial} compare different pairwise spatial features defined in~(\ref{eqn:pairwise_spatial_features}).
R4 only uses pairwise distances, while R5 only uses pairwise orientations.
We can see that pairwise distances are beneficial for view-independent sentences which contain distance-related spatial relations such as \emph{``next to''} and \emph{``farthest from''}, but do not improve the performance for view-dependent sentences.
In contrast, pairwise orientations significantly boost the performance for view-dependent sentences (+10.2\%). 
This shows that explicitly encoding the relative orientations facilitates the learning of view dependent spatial relations such as \emph{``in front of''} and \emph{``to the left of''}.
The relative distance and orientation features are also complementary. Their combination achieves the best performance as shown in our full model (R9).

\begin{wraptable}{l}{0.3\linewidth}
\centering
\caption{Comparison of pairwise spatial feature computation methods.}
\label{tab:ablation_nr3d_spatial_features}
\begin{tabular}{cc} \toprule
 & Overall \\ \midrule
object center & 74.4 \\
bottom center & 74.4 \\
boxes + MLP & 57.4 \\ \bottomrule
\end{tabular}
\end{wraptable}
Table~\ref{tab:ablation_nr3d_spatial_breakdown} provides a more systematic analysis of R3-5 and R9 in Table~\ref{tab:ablation_nr3d_spatial} to show the contribution of spatial relation modeling to different types of sentences.
We categorize sentences into four groups according to whether the sentence describes spatial relations in terms of distances or orientations.
We can see that the explicit pairwise distance modeling contributes most to the distance-only sentences but has little influence on samples with orientation-related sentences. On the other hand, the pairwise orientation modeling can significantly improve orientation-related sentences by 10.5\%. Combining both pairwise distance and orientation modeling achieves the best performance on all categories.

\begin{table}
\centering
\caption{Performance breakdown of models in Table~\ref{tab:ablation_nr3d_spatial} by spatial relation types: \emph{Dist(Ort) only} which only contains distance(orientation) descriptions; \emph{Dist \& Ort} which contains both distance and orientation descriptions; and the \emph{Others} which do not contain spatial relation descriptions.}
\label{tab:ablation_nr3d_spatial_breakdown}
\begin{tabular}{ccccccc} \toprule
Dist & Ort & Overall & \emph{Dist only} & \emph{Ort only} & \emph{Dist \& Ort} & \emph{Others} \\ \midrule
$\times$ & $\times$ & 62.4 & 63.5 & 61.2 & 57.7 & 63.9 \\
\checkmark & $\times$ & 66.0 & 72.6 & 58.9 & 55.1 & 68.8 \\
$\times$ & \checkmark & 71.3 & 73.8 & 71.7 & 67.7 & 69.1 \\
\checkmark & \checkmark & \textbf{74.4} & \textbf{77.8} & \textbf{74.0} & \textbf{69.1} & \textbf{72.6} \\ \bottomrule
\end{tabular}
\end{table}

\noindent\textbf{Pairwise spatial feature computation.}
We compute pairwise distances and angles using the coordinates of object centers and design the spatial features as shown in Eq~(\ref{eqn:pairwise_spatial_features}).
In Table~\ref{tab:ablation_nr3d_spatial_features}, we further compare the proposed method with two variants to show its effectiveness.
The first variant uses the bottom center of objects to compute pairwise vertical angles which ignores the height of objects and could be more accurate to measure vertical relations.
The second variant concatenates bounding boxes of two objects and uses a multi-layer perceptron (MLP) to learn pairwise spatial relations. 
We can see that using the bottom center achieves similar performance as using the object center, which suggests that the object center is sufficient to capture the vertical spatial relations of objects in the textual descriptions.
The learnable MLP, however, achieves much worse performance than our approach. This indicates that it is challenging to implicitly learn pairwise spatial relations and the proposed spatial features designed by domain knowledge are beneficial.

\noindent\textbf{Multi-head spatial attention.}
R6 in Table~\ref{tab:ablation_nr3d_spatial} uses single-head attention for the spatial relevance $\omega_{ij}^s$ in~(\ref{eqn:spatial_self_attn}) but keeps multi-head for the standard attention weight $\omega_{ij}^o$.
Compared to the full model R9 with the multi-head spatial attention, such single-head spatial attention achieves significantly worse performance.
This suggests that the multi-head is beneficial for learning spatial relations.

\begin{figure}[t]
    \centering
    \includegraphics[width=\linewidth]{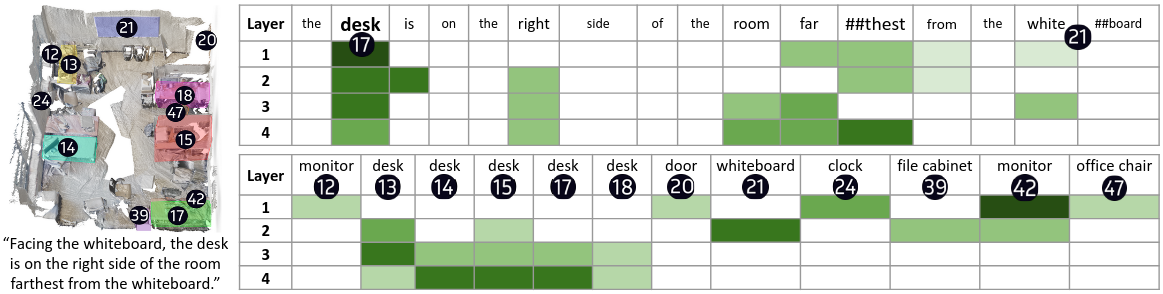}
    \caption{An example of the learned cross-attention (upper) and self-attention (bottom) at different transformer layers in the teacher model. The query is the target object proposal (desk 17). The darker color denotes higher attention weights from the query to the keys.}
    \label{fig:attn_viz}
    \vspace{-1em}
\end{figure}

\begin{wraptable}{r}{0.4\linewidth}
\centering
\caption{Performance comparison of different textual encoders.}
\label{tab:ablation_nr3d_text_encoder}
\begin{tabular}{ccc} \toprule
Encoder & \#Layers & Overall \\ \midrule
\multirow{4}{*}{BERT} & 3 & \textbf{74.4} \\
 & 6 & 73.7 \\
 & 9 & 73.2 \\
 & 12 & 52.3 \\ \midrule
Glove+GRU & 3 & 45.7 \\ \bottomrule
\end{tabular}
\end{wraptable}
\noindent\textbf{Attention fusion.}
We use sigmoid softmax function in~(\ref{eqn:spatial_self_attn}) to aggregate the spatial attention and the standard self-attention weights.
We compare the proposed fusion mechanism with standard relative positional encoding methods \cite{wu2021rethinking} in vision transformers: bias mode and contextual mode.
The bias mode adds the spatial attention weight as a bias term in the standard self-attention, while the contextual mode further considers the interaction with queries by injecting the pairwise spatial features into the key embeddings.
From the results in R7 and R8, we can see that the standard relative positional encoding methods fail to model 3D spatial relations among objects; they even perform worse than the strong baseline (R3), which doesn't rely on relative positional encoding.
The proposed sigmoid softmax function is by far more effective to modulate the standard self-attention with spatial relations. 

\noindent\textbf{Text encoding.}
We follow recent works \cite{yang2021sat,huang2022multi} to use the first three layers of BERT textual encoder, while \cite{yuan2021instancerefer} uses weaker textual encoder (Glove+GRU).
To compare these different textual encoders, Table~\ref{tab:ablation_nr3d_text_encoder} provides results of our proposed models under the setting in Table~\ref{tab:ablation_nr3d_spatial}. 
We can see that more BERT layers do not lead to better performance. The reason might be that the representations from higher BERT layers are less generalizable to different domains. Moreover, more layers are more prone to overfit and harder to optimize.
The pretrained BERT model achieves much better performance compared to the GRU model trained from scratch.

\subsubsection{Teacher-Student Training}

\begin{wraptable}{r}{0.46\linewidth}
    \vspace{-2em}
	\centering
	\caption{Grounding accuracy (\%) on the Nr3D dataset with ground-truth object proposals.}
	\label{tab:ablation_nr3d_training}
	\begin{tabular}{ccccc} \toprule
		& init. & $\mathcal{L}_{attn}$ & $\mathcal{L}_{hidden}$ & Overall \\ \midrule
		Teacher & \multicolumn{3}{c}{-} & 74.4 \\ \midrule
		\multirow{5}{*}{Student} & $\times$ & $\times$ & $\times$ & 58.1 \\
		& \checkmark & $\times$ & $\times$  & 62.6 \\
		 & $\times$ & \checkmark & $\times$ & 63.6 \\
		& $\times$ & $\times$ & \checkmark  & 62.1 \\
		& $\times$ & \checkmark & \checkmark  & \textbf{64.4} \\ \bottomrule
	\end{tabular}
\end{wraptable}

\begin{table}[t]
	\centering
	\tabcolsep=0.15cm
	\caption{Grounding accuracy (\%) on Nr3D and Sr3D datasets with ground-truth object proposals.}
	\label{tab:nr3d_sr3d_sota}
	\begin{tabular}{lcccccccccc} \toprule
		\multirow{3}{*}{Method} & \multicolumn{5}{c}{Nr3D} & \multicolumn{5}{c}{Sr3D}  \\ \cmidrule{2-11}
		 & Overall & Easy & Hard & \multicolumn{1}{c}{\begin{tabular}[c]{@{}c@{}}View\\ Dep\end{tabular}} & \multicolumn{1}{c}{\begin{tabular}[c]{@{}c@{}}View\\ Indep\end{tabular}} & Overall & Easy & Hard & \multicolumn{1}{c}{\begin{tabular}[c]{@{}c@{}}View\\ Dep\end{tabular}} & \multicolumn{1}{c}{\begin{tabular}[c]{@{}c@{}}View\\ Indep\end{tabular}}  \\ \midrule
		ReferIt3D \cite{achlioptas2020referit3d} & 35.6 & 43.6 & 27.9 & 32.5 & 37.1 & 40.8 & 44.7 & 31.5 & 39.2 & 40.8 \\
		ScanRefer \cite{chen2020scanrefer} & 34.2 & 41.0 & 23.5 & 29.9 & 35.4 & - & - & - & - & -\\
		TGNN \cite{huang2021text}  & 37.3 & 44.2 & 30.6 & 35.8 & 38.0 & - & - & - & - & - \\
		InstanceRefer \cite{yuan2021instancerefer} & 38.8 & 46.0 & 31.8 & 34.5 & 41.9 & 48.0 & 51.1 & 40.5 & 45.4 & 48.1 \\
		FFL-3DOG \cite{feng2021free} & 41.7 & 48.2 & 35.0 & 37.1 & 44.7 & - & - & - & - & - \\
		3DVG-Trans \cite{zhao20213dvg} & 40.8 & 48.5 & 34.8 & 34.8 & 43.7 & 51.4 & 54.2 & 44.9 & 44.6 & 51.7 \\
		TransRefer3D \cite{he2021transrefer3d}  & 42.1 & 48.5 & 36.0 & 36.5 & 44.9 & 57.4 & 60.5 & 50.2 & 49.9 & 57.7 \\
		LanguageRefer \cite{roh2022languagerefer}  & 43.9 & 51.0 & 36.6 & 41.7 & 45.0 & 56.0 & 58.9 & 49.3 & 49.2 & 56.3 \\
		SAT \cite{yang2021sat} & 49.2 & 56.3 & 42.4 & 46.9 & 50.4 & 57.9 & 61.2 & 50.0 & 49.2 & 58.3 \\
		3D-SPS \cite{luo20223dsps} & 51.5 & 58.1 & 45.1 & 48.0 & 53.2 & 62.6 & 56.2 & 65.4 & 49.2 & 63.2 \\
		Multi-view \cite{huang2022multi} & 55.1 & 61.3 & 49.1 & 54.3 & 55.4 & 64.5 & 66.9 & 58.8 & 58.4 & 64.7 \\ \midrule
		\modelname~(Ours) & \textbf{64.4} & \textbf{70.2} & \textbf{57.4}  & \textbf{62.0}  & \textbf{64.5} & \textbf{72.8} & \textbf{74.9} & \textbf{67.9}  & \textbf{63.8}  & \textbf{73.2}  \\ \bottomrule
	\end{tabular}
\end{table}

The teacher model with ground-truth object labels learns well how to perform spatial relation reasoning given the input sentence.
Figure~\ref{fig:attn_viz} provides a qualitative example of the learned cross- and self-attention weights at all multimodal fusion layers in the teacher model.
For the cross-attention weights, we can see that the teacher first aligns the object proposal with the correct words in the text, and then gradually shifts its attentions to relation words to perform spatial relation reasoning.
For the self-attention weights, in the first layer, the teacher mostly attends to nearby objects to be aware of the context; then in the second layer, it focuses more on the reference object mentioned in the sentence (\textit{e.g.}, the whiteboard); finally, it concentrates on the same-class distractors to distinguish them.
Such reasoning steps are promising to ease the training of the student model.
In the following, we evaluate the proposed teacher-student training to distill relational knowledge from the teacher to a student with point cloud inputs extracted from the ground-truth object proposals. 

Table~\ref{tab:ablation_nr3d_training} presents the performance of different student models.
The first row in the student block does not use any knowledge from the teacher model.
In the second row, we use the weights in the teacher model to initialize the student model since the teacher and student share the same architecture except for the input object representations.
Such simple weight initialization already facilitates the training of the student, and achieves 4.5\% gains.
The attention distillation in~(\ref{eqn:attn_distill_loss}) and hidden state distillation in~(\ref{eqn:hidden_distill_loss}) are both beneficial to improve the student model compared to the first row without knowledge distillation.
It is more effective to distill the relation knowledge in the attention weight matrices.
The combination of $\mathcal{L}_{attn}$ and $\mathcal{L}_{hidden}$ is helpful and achieves the best performance. 
We empirically find that combining weight initialization and the two distillation losses performs similar to using the distillation losses alone.

\subsection{Comparison with State-of-the-Art Methods}

Table~\ref{tab:nr3d_sr3d_sota} compares our \modelname~model with state-of-the-art methods on Nr3D and Sr3D datasets.
All the compared works use ground-truth object proposals, but no ground-truth labels. 
Our model achieves significant improvements over the previous best method \cite{huang2022multi}, with 9.3\% and 8.3\% absolute gains on Nr3D and Sr3D respectively.
We outperform \cite{zhao20213dvg} by even larger margins on the two datasets. This work also use  spatial information, but doesn't use a spatial attention module.

Table~\ref{tab:scanrefer_sota_gt} and Table~\ref{tab:scanrefer_sota_det} present results on ScanRefer dataset with ground-truth object proposals and detected object proposals, respectively.
For fair comparison, we use a pre-trained PointGroup \cite{jiang2020pointgroup} model to generate object proposals, which is trained on 18 classes in ScanNet.
To be noted, the upper block in Table~\ref{tab:scanrefer_sota_det} optimizes the detection stage or utilizes single-stage pipeline.
These methods use 
\begin{wraptable}{r}{0.32\linewidth}
\vspace{-1em}
\centering
\caption{Grounding accuracy (\%) on ScanRefer with ground-truth object proposals.}
\label{tab:scanrefer_sota_gt}
\begin{tabular}{cc} \toprule
 & Overall \\ \midrule
ReferIt3D \cite{achlioptas2020referit3d} & 46.9 \\
Non-SAT \cite{yang2021sat} & 48.2 \\
SAT \cite{yang2021sat} & 53.8 \\ \midrule
\modelname~(Ours) & \textbf{59.8} \\ \bottomrule
\end{tabular}
\end{wraptable}
more data to train object proposals, so the comparison to these methods is not entirely fair. 
There is a large gap between the ground-truth and detected proposals (over 12\% for our method), because the detected proposals might not include the target or reference objects in the sentence.
However, our \modelname~model still outperforms the state of the art in both settings.
Though our proposed spatial relation module can also be plugged into transformer-based 3D object detectors \cite{liu2021group} to improve the 3D object proposal generation, we leave it to future work.
Figure~\ref{fig:intro} compares our \modelname~model and a baseline without spatial self-attention and teacher-student training.
We present more qualitative comparisons in Section~\ref{sec:supp_examples} of supplementary material.

\begin{table}
\centering
\tabcolsep=0.15cm
\caption{Grounding accuracy (\%) on ScanRefer with detected object proposals. VN and PG denote the VoteNet and PointGroup models pretrained on 18 classes of the ScanNet dataset, while Optimized denotes end-to-end training of the object detector on the ScanRefer dataset.}
\label{tab:scanrefer_sota_det}
\begin{tabular}{lccccccc} \toprule
\multirow{2}{*}{Method} & \multirow{2}{*}{Det} & \multicolumn{2}{c}{Unique} & \multicolumn{2}{c}{Multiple} & \multicolumn{2}{c}{Overall} \\ 
 &  & acc@0.25 & acc@0.5 & acc@0.25 & acc@0.5 & acc@0.25 & acc@0.5 \\ \midrule
ScanRefer \cite{chen2020scanrefer} & \multirow{4}{*}{\begin{tabular}[c]{@{}c@{}}Opti-\\ mized\end{tabular}} & 65.00 & 43.31 & 30.63 & 19.75 & 37.30 & 24.32 \\
TGNN \cite{huang2021text} &  & 68.61 & 56.80 & 29.84 & 23.18 & 37.37 & 29.70 \\
3DVG-Trans \cite{zhao20213dvg} &  & 77.16 & 58.47 & 38.38 & 28.70 & 45.90 & 34.47 \\
3D-SPS \cite{luo20223dsps} & & 81.63 & 64.77 & 39.48 & 29.61 & 47.65 & 36.42 \\ \midrule
Non-SAT \cite{yang2021sat} & VN \cite{qi2019deep} & 68.48 & 47.38 & 31.81 & 21.34 & 38.92 & 26.40 \\
SAT \cite{yang2021sat} & VN \cite{qi2019deep} & 73.21 & 50.83 & 37.64 & 25.16 & 44.54 & 30.14 \\
InstanceRefer \cite{yuan2021instancerefer} & PG \cite{jiang2020pointgroup} & 77.45 & 66.83 & 31.27 & 24.77 & 40.23 & 32.93 \\
Multi-view \cite{huang2022multi} & PG \cite{jiang2020pointgroup} & 77.67 & 66.45 & 31.92 & 25.26 & 40.80 & 33.26 \\
\modelname~(Ours) & PG \cite{jiang2020pointgroup} & \textbf{81.58} &  \textbf{68.62} & \textbf{40.30} & \textbf{30.71} & \textbf{47.94} & \textbf{37.73} \\ \midrule
UpperBound & PG \cite{jiang2020pointgroup} & 88.63 & 74.47 & 78.82 & 60.37 & 80.64 & 62.98 \\ \bottomrule 

\end{tabular}
\vspace{-1em}
\end{table}

\section{Conclusion}
In this work, we propose a \modelname~model for 3D object grounding.
It contains a newly designed spatial self-attention module to improve language conditioned spatial relation reasoning in the transformer layer, which explicitly considers relative distances and orientations among objects.
A teacher-student training approach is further proposed to transfer relation knowledge from a teacher with ground-truth object labels to a student with point cloud inputs.
The proposed model significantly outperforms the state of the art on Nr3D, Sr3D and ScanRefer datasets. 

Beyond the application of 3D object grounding, the general approach of incorporating priors into a self-attention layer for a transformer to combine multimodal input can be of wide interest.
Since our model belongs to the two-stage framework, it is limited by imperfect object proposals in the first detection stage.
We also did not explore to explicitly extract object orientations to more accurately estimate pairwise spatial relations as the automatic object poses prediction remains a challenging problem.
In addition, the evaluation can suffer from the fact that existing datasets do not represent a rich diversity of environments though we carried out extensive ablations to mitigate the problem.
This work has minimal ethical, privacy and safety concerns.

\begin{ack}
This work was granted access to the HPC resources of IDRIS under the allocation 101002 made by GENCI. 
It was funded in part by the French government under management of Agence Nationale de la Recherche as part of the “Investissements d’avenir” program, reference ANR19-P3IA-0001 (PRAIRIE 3IA Institute), the ANR project VideoPredict (ANR-21-FAI1-0002-01) and by Louis Vuitton ENS Chair on Artificial Intelligence.
\end{ack}


\begin{thebibliography}{10}

\bibitem{yu2016modeling}
Licheng Yu, Patrick Poirson, Shan Yang, Alexander~C Berg, and Tamara~L Berg.
\newblock Modeling context in referring expressions.
\newblock In {\em ECCV}, pages 69--85. Springer, 2016.

\bibitem{mao2016generation}
Junhua Mao, Jonathan Huang, Alexander Toshev, Oana Camburu, Alan~L Yuille, and
  Kevin Murphy.
\newblock Generation and comprehension of unambiguous object descriptions.
\newblock In {\em CVPR}, pages 11--20, 2016.

\bibitem{liu2019clevr}
Runtao Liu, Chenxi Liu, Yutong Bai, and Alan~L Yuille.
\newblock Clevr-ref+: Diagnosing visual reasoning with referring expressions.
\newblock In {\em CVPR}, pages 4185--4194, 2019.

\bibitem{chen2020cops}
Zhenfang Chen, Peng Wang, Lin Ma, Kwan-Yee~K Wong, and Qi~Wu.
\newblock Cops-ref: A new dataset and task on compositional referring
  expression comprehension.
\newblock In {\em CVPR}, pages 10086--10095, 2020.

\bibitem{chen2019touchdown}
Howard Chen, Alane Suhr, Dipendra Misra, Noah Snavely, and Yoav Artzi.
\newblock Touchdown: Natural language navigation and spatial reasoning in
  visual street environments.
\newblock In {\em CVPR}, pages 12538--12547, 2019.

\bibitem{qi2020reverie}
Yuankai Qi, Qi~Wu, Peter Anderson, Xin Wang, William~Yang Wang, Chunhua Shen,
  and Anton van~den Hengel.
\newblock Reverie: Remote embodied visual referring expression in real indoor
  environments.
\newblock In {\em CVPR}, pages 9982--9991, 2020.

\bibitem{chen2020scanrefer}
Dave~Zhenyu Chen, Angel~X Chang, and Matthias Nie{\ss}ner.
\newblock Scanrefer: 3d object localization in rgb-d scans using natural
  language.
\newblock In {\em ECCV}, pages 202--221. Springer, 2020.

\bibitem{achlioptas2020referit3d}
Panos Achlioptas, Ahmed Abdelreheem, Fei Xia, Mohamed Elhoseiny, and Leonidas
  Guibas.
\newblock Referit3d: Neural listeners for fine-grained 3d object identification
  in real-world scenes.
\newblock In {\em ECCV}, pages 422--440. Springer, 2020.

\bibitem{he2021transrefer3d}
Dailan He, Yusheng Zhao, Junyu Luo, Tianrui Hui, Shaofei Huang, Aixi Zhang, and
  Si~Liu.
\newblock Transrefer3d: Entity-and-relation aware transformer for fine-grained
  3d visual grounding.
\newblock In {\em ACM MM}, pages 2344--2352, 2021.

\bibitem{huang2021text}
Pin-Hao Huang, Han-Hung Lee, Hwann-Tzong Chen, and Tyng-Luh Liu.
\newblock Text-guided graph neural networks for referring 3d instance
  segmentation.
\newblock In {\em AAAI}, volume~35, pages 1610--1618, 2021.

\bibitem{yuan2021instancerefer}
Zhihao Yuan, Xu~Yan, Yinghong Liao, Ruimao Zhang, Sheng Wang, Zhen Li, and
  Shuguang Cui.
\newblock Instancerefer: Cooperative holistic understanding for visual
  grounding on point clouds through instance multi-level contextual referring.
\newblock In {\em ICCV}, pages 1791--1800, 2021.

\bibitem{feng2021free}
Mingtao Feng, Zhen Li, Qi~Li, Liang Zhang, XiangDong Zhang, Guangming Zhu, Hui
  Zhang, Yaonan Wang, and Ajmal Mian.
\newblock Free-form description guided 3d visual graph network for object
  grounding in point cloud.
\newblock In {\em ICCV}, pages 3722--3731, 2021.

\bibitem{zhao20213dvg}
Lichen Zhao, Daigang Cai, Lu~Sheng, and Dong Xu.
\newblock 3dvg-transformer: Relation modeling for visual grounding on point
  clouds.
\newblock In {\em ICCV}, pages 2928--2937, 2021.

\bibitem{yang2021sat}
Zhengyuan Yang, Songyang Zhang, Liwei Wang, and Jiebo Luo.
\newblock Sat: 2d semantics assisted training for 3d visual grounding.
\newblock In {\em ICCV}, pages 1856--1866, 2021.

\bibitem{roh2022languagerefer}
Junha Roh, Karthik Desingh, Ali Farhadi, and Dieter Fox.
\newblock Languagerefer: Spatial-language model for 3d visual grounding.
\newblock In {\em CoRL}, pages 1046--1056. PMLR, 2021.

\bibitem{huang2022multi}
Shijia Huang, Yilun Chen, Jiaya Jia, and Liwei Wang.
\newblock Multi-view transformer for 3d visual grounding.
\newblock In {\em CVPR}, 2022.

\bibitem{vaswani2017attention}
Ashish Vaswani, Noam Shazeer, Niki Parmar, Jakob Uszkoreit, Llion Jones,
  Aidan~N Gomez, {\L}ukasz Kaiser, and Illia Polosukhin.
\newblock Attention is all you need.
\newblock {\em NeurIPS}, 30, 2017.

\bibitem{kamath2021mdetr}
Aishwarya Kamath, Mannat Singh, Yann LeCun, Gabriel Synnaeve, Ishan Misra, and
  Nicolas Carion.
\newblock Mdetr-modulated detection for end-to-end multi-modal understanding.
\newblock In {\em ICCV}, pages 1780--1790, 2021.

\bibitem{subramanian2022reclip}
Sanjay Subramanian, Will Merrill, Trevor Darrell, Matt Gardner, Sameer Singh,
  and Anna Rohrbach.
\newblock Reclip: A strong zero-shot baseline for referring expression
  comprehension.
\newblock In {\em ACL}, 2022.

\bibitem{krishna2017visual}
Ranjay Krishna, Yuke Zhu, Oliver Groth, Justin Johnson, Kenji Hata, Joshua
  Kravitz, Stephanie Chen, Yannis Kalantidis, Li-Jia Li, David~A Shamma, et~al.
\newblock Visual genome: Connecting language and vision using crowdsourced
  dense image annotations.
\newblock {\em IJCV}, 123(1):32--73, 2017.

\bibitem{radford2021learning}
Alec Radford, Jong~Wook Kim, Chris Hallacy, Aditya Ramesh, Gabriel Goh,
  Sandhini Agarwal, Girish Sastry, Amanda Askell, Pamela Mishkin, Jack Clark,
  et~al.
\newblock Learning transferable visual models from natural language
  supervision.
\newblock In {\em ICML}, pages 8748--8763. PMLR, 2021.

\bibitem{projectpage}
Project webpage.
\newblock \url{https://cshizhe.github.io/projects/vil3dref.html}.

\bibitem{luo20223dsps}
Junyu Luo, Jiahui Fu, Xianghao Kong, Chen Gao, Haibing Ren, Hao Shen, Huaxia
  Xia, and Si~Liu.
\newblock 3d-sps: Single-stage 3d visual grounding via referred point
  progressive selection.
\newblock In {\em CVPR}, 2022.

\bibitem{yu2018mattnet}
Licheng Yu, Zhe Lin, Xiaohui Shen, Jimei Yang, Xin Lu, Mohit Bansal, and
  Tamara~L Berg.
\newblock Mattnet: Modular attention network for referring expression
  comprehension.
\newblock In {\em Proceedings of the IEEE Conference on Computer Vision and
  Pattern Recognition}, pages 1307--1315, 2018.

\bibitem{dai2017scannet}
Angela Dai, Angel~X Chang, Manolis Savva, Maciej Halber, Thomas Funkhouser, and
  Matthias Nie{\ss}ner.
\newblock Scannet: Richly-annotated 3d reconstructions of indoor scenes.
\newblock In {\em Proceedings of the IEEE conference on computer vision and
  pattern recognition}, pages 5828--5839, 2017.

\bibitem{velivckovic2018graph}
Petar Veli{\v{c}}kovi{\'c}, Guillem Cucurull, Arantxa Casanova, Adriana Romero,
  Pietro Lio, and Yoshua Bengio.
\newblock Graph attention networks.
\newblock {\em ICLR}, 2018.

\bibitem{wang2019dynamic}
Yue Wang, Yongbin Sun, Ziwei Liu, Sanjay~E Sarma, Michael~M Bronstein, and
  Justin~M Solomon.
\newblock Dynamic graph cnn for learning on point clouds.
\newblock {\em TOG}, 38(5):1--12, 2019.

\bibitem{jain2021looking}
Ayush Jain, Nikolaos Gkanatsios, Ishita Mediratta, and Katerina Fragkiadaki.
\newblock Looking outside the box to ground language in 3d scenes.
\newblock {\em arXiv preprint arXiv:2112.08879}, 2021.

\bibitem{sun2019videobert}
Chen Sun, Austin Myers, Carl Vondrick, Kevin Murphy, and Cordelia Schmid.
\newblock Videobert: A joint model for video and language representation
  learning.
\newblock In {\em ICCV}, pages 7464--7473, 2019.

\bibitem{yang2022tubedetr}
Antoine Yang, Antoine Miech, Josef Sivic, Ivan Laptev, and Cordelia Schmid.
\newblock Tubedetr: Spatio-temporal video grounding with transformers.
\newblock {\em CVPR}, 2022.

\bibitem{cornia2020meshed}
Marcella Cornia, Matteo Stefanini, Lorenzo Baraldi, and Rita Cucchiara.
\newblock Meshed-memory transformer for image captioning.
\newblock In {\em CVPR}, pages 10578--10587, 2020.

\bibitem{zhou2020unified}
Luowei Zhou, Hamid Palangi, Lei Zhang, Houdong Hu, Jason Corso, and Jianfeng
  Gao.
\newblock Unified vision-language pre-training for image captioning and vqa.
\newblock In {\em AAAI}, volume~34, pages 13041--13049, 2020.

\bibitem{chen2021history}
Shizhe Chen, Pierre-Louis Guhur, Cordelia Schmid, and Ivan Laptev.
\newblock History aware multimodal transformer for vision-and-language
  navigation.
\newblock {\em NeurIPS}, 34, 2021.

\bibitem{chen2022think}
Shizhe Chen, Pierre-Louis Guhur, Makarand Tapaswi, Cordelia Schmid, and Ivan
  Laptev.
\newblock Think global, act local: Dual-scale graph transformer for
  vision-and-language navigation.
\newblock {\em CVPR}, 2022.

\bibitem{wu2021rethinking}
Kan Wu, Houwen Peng, Minghao Chen, Jianlong Fu, and Hongyang Chao.
\newblock Rethinking and improving relative position encoding for vision
  transformer.
\newblock In {\em ICCV}, pages 10033--10041, 2021.

\bibitem{hinton2015distilling}
Geoffrey Hinton, Oriol Vinyals, Jeff Dean, et~al.
\newblock Distilling the knowledge in a neural network.
\newblock {\em arXiv preprint arXiv:1503.02531}, 2(7), 2015.

\bibitem{beyer2021knowledge}
Lucas Beyer, Xiaohua Zhai, Am{\'e}lie Royer, Larisa Markeeva, Rohan Anil, and
  Alexander Kolesnikov.
\newblock Knowledge distillation: A good teacher is patient and consistent.
\newblock {\em arXiv preprint arXiv:2106.05237}, 2021.

\bibitem{jiao2020tinybert}
Xiaoqi Jiao, Yichun Yin, Lifeng Shang, Xin Jiang, Xiao Chen, Linlin Li, Fang
  Wang, and Qun Liu.
\newblock Tinybert: Distilling bert for natural language understanding.
\newblock In {\em EMNLP Findings}, pages 4163--4174, 2020.

\bibitem{devlin2019bert}
Jacob Devlin, Ming-Wei Chang, Kenton Lee, and Kristina Toutanova.
\newblock Bert: Pre-training of deep bidirectional transformers for language
  understanding.
\newblock {\em NAACL}, 2019.

\bibitem{qi2017pointnet++}
Charles~Ruizhongtai Qi, Li~Yi, Hao Su, and Leonidas~J Guibas.
\newblock Pointnet++: Deep hierarchical feature learning on point sets in a
  metric space.
\newblock {\em NeurIPS}, 30, 2017.

\bibitem{pennington2014glove}
Jeffrey Pennington, Richard Socher, and Christopher~D Manning.
\newblock Glove: Global vectors for word representation.
\newblock In {\em Proceedings of the 2014 conference on empirical methods in
  natural language processing (EMNLP)}, pages 1532--1543, 2014.

\bibitem{loshchilov2018fixing}
Ilya Loshchilov and Frank Hutter.
\newblock Fixing weight decay regularization in adam.
\newblock 2018.

\bibitem{jiang2020pointgroup}
Li~Jiang, Hengshuang Zhao, Shaoshuai Shi, Shu Liu, Chi-Wing Fu, and Jiaya Jia.
\newblock Pointgroup: Dual-set point grouping for 3d instance segmentation.
\newblock In {\em CVPR}, pages 4867--4876, 2020.

\bibitem{liu2021group}
Ze~Liu, Zheng Zhang, Yue Cao, Han Hu, and Xin Tong.
\newblock Group-free 3d object detection via transformers.
\newblock In {\em Proceedings of the IEEE/CVF International Conference on
  Computer Vision}, pages 2949--2958, 2021.

\bibitem{qi2019deep}
Charles~R Qi, Or~Litany, Kaiming He, and Leonidas~J Guibas.
\newblock Deep hough voting for 3d object detection in point clouds.
\newblock In {\em ICCV}, pages 9277--9286, 2019.

\end{thebibliography}

\section*{Checklist}

\begin{enumerate}

\item For all authors...
\begin{enumerate}
  \item Do the main claims made in the abstract and introduction accurately reflect the paper's contributions and scope?
    \answerYes{}
  \item Did you describe the limitations of your work?
    \answerYes{See the conclusion section.}
  \item Did you discuss any potential negative societal impacts of your work?
    \answerYes{See the conclusion section.}
  \item Have you read the ethics review guidelines and ensured that your paper conforms to them?
    \answerYes{}
\end{enumerate}

\item If you are including theoretical results...
\begin{enumerate}
  \item Did you state the full set of assumptions of all theoretical results?
    \answerNA{}
        \item Did you include complete proofs of all theoretical results?
    \answerNA{}
\end{enumerate}

\item If you ran experiments...
\begin{enumerate}
  \item Did you include the code, data, and instructions needed to reproduce the main experimental results (either in the supplemental material or as a URL)?
    \answerNo{We will release the code, data and instructions upon acceptance.}
  \item Did you specify all the training details (e.g., data splits, hyperparameters, how they were chosen)?
    \answerYes{See the experiment section.}
        \item Did you report error bars (e.g., with respect to the random seed after running experiments multiple times)?
    \answerYes{See the supplementary material.}
        \item Did you include the total amount of compute and the type of resources used (e.g., type of GPUs, internal cluster, or cloud provider)?
    \answerYes{See the experimental section.}
\end{enumerate}

\item If you are using existing assets (e.g., code, data, models) or curating/releasing new assets...
\begin{enumerate}
  \item If your work uses existing assets, did you cite the creators?
    \answerYes{}
  \item Did you mention the license of the assets?
    \answerYes{}
  \item Did you include any new assets either in the supplemental material or as a URL?
    \answerNo{}
  \item Did you discuss whether and how consent was obtained from people whose data you're using/curating?
    \answerNA{}
  \item Did you discuss whether the data you are using/curating contains personally identifiable information or offensive content?
    \answerNA{}
\end{enumerate}

\item If you used crowdsourcing or conducted research with human subjects...
\begin{enumerate}
  \item Did you include the full text of instructions given to participants and screenshots, if applicable?
    \answerNA{}
  \item Did you describe any potential participant risks, with links to Institutional Review Board (IRB) approvals, if applicable?
    \answerNA{}
  \item Did you include the estimated hourly wage paid to participants and the total amount spent on participant compensation?
    \answerNA{}
\end{enumerate}

\end{enumerate}

\appendix

\section*{Appendix}

\section{Implementation Details}
\label{sec:supp_model}

\subsection{Computing horizontal and vertical angles between two objects}
For two objects A and B, assume that the coordinates of the object center are $(x_a, y_a, z_a)$ and $(x_b, y_b, z_b)$. We can compute the Euclidean distance between A and B as $d$. The horizontal angle $\theta_h$ is $\text{arctan2}(\frac{y_b-y_a}{x_b-x_a})$ and the vertical angle $\theta_v$ is $\text{arcsin}(\frac{z_b-z_a}{d})$. 
In other terminology, the horizontal angle corresponds to the azimuth direction while looking from point A to B, while the vertical angle corresponds to the elevation.

\subsection{Rotation augmentation}
When performing rotation augmentation, we rotate the whole point cloud by different angles. Specifically, we randomly select one angle among [0, 90, 180, 270] degrees.

\subsection{Training losses}
We use auxiliary losses $L_{sent}$ and $L^*_{obj}$ following previous works~\cite{chen2020scanrefer,he2021transrefer3d,zhao20213dvg,yang2021sat,huang2022multi} which have shown to be beneficial for the performance.
They are all cross entropy losses. 
$L_{sent}$ is to predict the target object class from the sentence. 
$L^*_{obj}$ is to predict the object class for each object token.
When training the teacher model, all the losses are used.
When training the student model, the $L^u_{obj}$ does not influence model weights since the pointnet is fixed.

\section{Additional Results}
\label{sec:supp_results}

\subsection{Ablations Studies on ScanRefer Dataset}
In the main paper, we provide ablation studies on the Nr3D dataset. Here, we further evaluate our method on the ScanRefer dataset to demonstrate the effectiveness of the proposed spatial self-attention and teacher-student training.

\begin{wraptable}{r}{0.7\linewidth}
	\centering
	\caption{Grounding accuracy (\%) on the ScanRefer dataset with ground-truth object labels. Dist stands for Distance; Ort for Orientation; MHA for Multi-Head spatial Attention; RotAug for Rotation Augmentation; sigs for sigsoftmax in Eq~(5); and `-' means not applicable.}
	\label{tab:scanrefer_ablation_gtlabels}
	\begin{tabular}{cccccccccc} \toprule
		& \multicolumn{4}{c}{Spatial Relation Reasoning} & \multirow{2}{*}{\begin{tabular}[c]{@{}c@{}}Rot\\ Aug\end{tabular}} & \multirow{2}{*}{Color} & \multirow{2}{*}{Overall} \\
		& Dist & Ort & MHA & Fusion &  &  &  \\ \midrule
		R1 & - & - & - & - & $\times$ & $\times$ & 56.35 \\
		R2 & - & - & - & - & $\times$ & \checkmark & 57.80 \\
		R3 & - & - & - & - & \checkmark & \checkmark & 59.10 \\ \midrule
		R4 & \checkmark & $\times$ & \checkmark & sigs & \checkmark & \checkmark &  63.87 \\
		R5 & $\times$ & \checkmark & \checkmark & sigs & \checkmark & \checkmark &  60.28 \\
		R6 & \checkmark & \checkmark & $\times$ & sigs & \checkmark & \checkmark &  63.39\\ \midrule
		R7 & \checkmark & \checkmark & \checkmark & bias & \checkmark & \checkmark & 59.16 \\
		R8 & \checkmark & \checkmark & \checkmark & ctx & \checkmark & \checkmark & 58.79 \\ \midrule
		R9 & \checkmark & \checkmark & \checkmark & sigs & \checkmark & \checkmark &  \textbf{63.89} \\ \bottomrule
	\end{tabular}
\end{wraptable}
Table~\ref{tab:scanrefer_ablation_gtlabels} presents the grounding accuracy using the ground-truth object labels on the ScanRefer dataset.
By comparing R1-R3, we observe that the color information and rotation augmentation is beneficial to provide a stronger baseline.
The proposed spatial self-attention (R9) significantly improves the strong baseline (R3), with the absolute gain of 4.79\%.
However, different from the Nr3D dataset, the sentences in the ScanRefer dataset are less focused on spatial relations and contain more attribute descriptions such as colors and shapes to refer to target objects.
Therefore, the overall performance gain from R3 to R9 is smaller compared to the Nr3D dataset.
The rows R4 and R5 analyze the contributions of relative spatial features to the performance. We can see that both relative distances and orientations outperform the baseline in R3, and that the relative distances are more important in the ScanRefer dataset. Their combination achieves the best performance in R9.
The multi-head attention is also beneficial comparing R6 and R9.
In R7 and R8, we compare the proposed sigmoid softmax function with the other two common relative position encoding methods. Our proposed fusion method is most effective to exploit both the spatial attention and the standard self-attention weights.

\begin{wraptable}{l}{0.5\linewidth}
	\centering
	\caption{Grounding accuracy (\%) on the ScanRefer dataset with ground-truth object proposals.}
	\label{tab:scanrefer_ablation_gtpcds}
	\begin{tabular}{ccccc} \toprule
		& init. & $\mathcal{L}_{attn}$ & $\mathcal{L}_{hidden}$ & Overall \\ \midrule
		Teacher & \multicolumn{3}{c}{-} & 63.89 \\ \midrule
		\multirow{5}{*}{Student} & $\times$ & $\times$ & $\times$ & 57.50 \\
		& \checkmark & $\times$ & $\times$  & 57.81 \\
		& $\times$ & \checkmark & $\times$ & 59.67 \\
		& $\times$ & $\times$ & \checkmark  & \textbf{59.89} \\
		& $\times$ & \checkmark & \checkmark  & 59.80 \\ \bottomrule
	\end{tabular}
\end{wraptable}

Table~\ref{tab:scanrefer_ablation_gtpcds} compares models using ground-truth object proposals on the ScanRefer dataset.
The first row in the student model block does not use any knowledge from the teacher model which uses ground-truth object labels as inputs.
Initializing from the weights in the teacher only brings a slight improvement as shown in the second row.
However, our proposed knowledge distillation achieves over 2\% boost compared to the baseline in the first row.
Both attention weights and hidden states are beneficial to train a better student model with noisy object features. The hidden state distillation slightly outperforms attention distillation which is different from the results on the Nr3D dataset. The reason could be that the spatial relations are mentioned more frequently in the Nr3D dataset while object attributes are more typical in the ScanRefer dataset.

\subsection{Robustness to Random Seeds}
\begin{wraptable}{r}{0.5\linewidth}
\vspace{-1em}
	\centering
	\caption{The robustness of the proposed ViL3DRel model with respect to different random seeds (\%).}
	\label{tab:expr_random_seeds}
	\begin{tabular}{cccc} \toprule
		& Nr3D & Sr3D & ScanRefer \\ \midrule
		seed=0 & 64.4 & 72.8 & 37.7 \\
		5 seeds & 63.8 $\pm$ 0.5 & 72.8 $\pm$ 0.2 & 37.6 $\pm$ 0.2 \\ \bottomrule
	\end{tabular}
\end{wraptable}
We use the same random seed of 0 for all the experiments.
In the following, we verify the robustness of the proposed ViL3DRel model by measuring the average and standard derivations of the grounding accuracy under 5 different random seeds.
From the results in Table~\ref{tab:expr_random_seeds}, we can see that the proposed model is stable with low deviations across different random seeds.

\section{Qualitative Results}
\label{sec:supp_examples}

\begin{figure}
    \includegraphics[width=\linewidth]{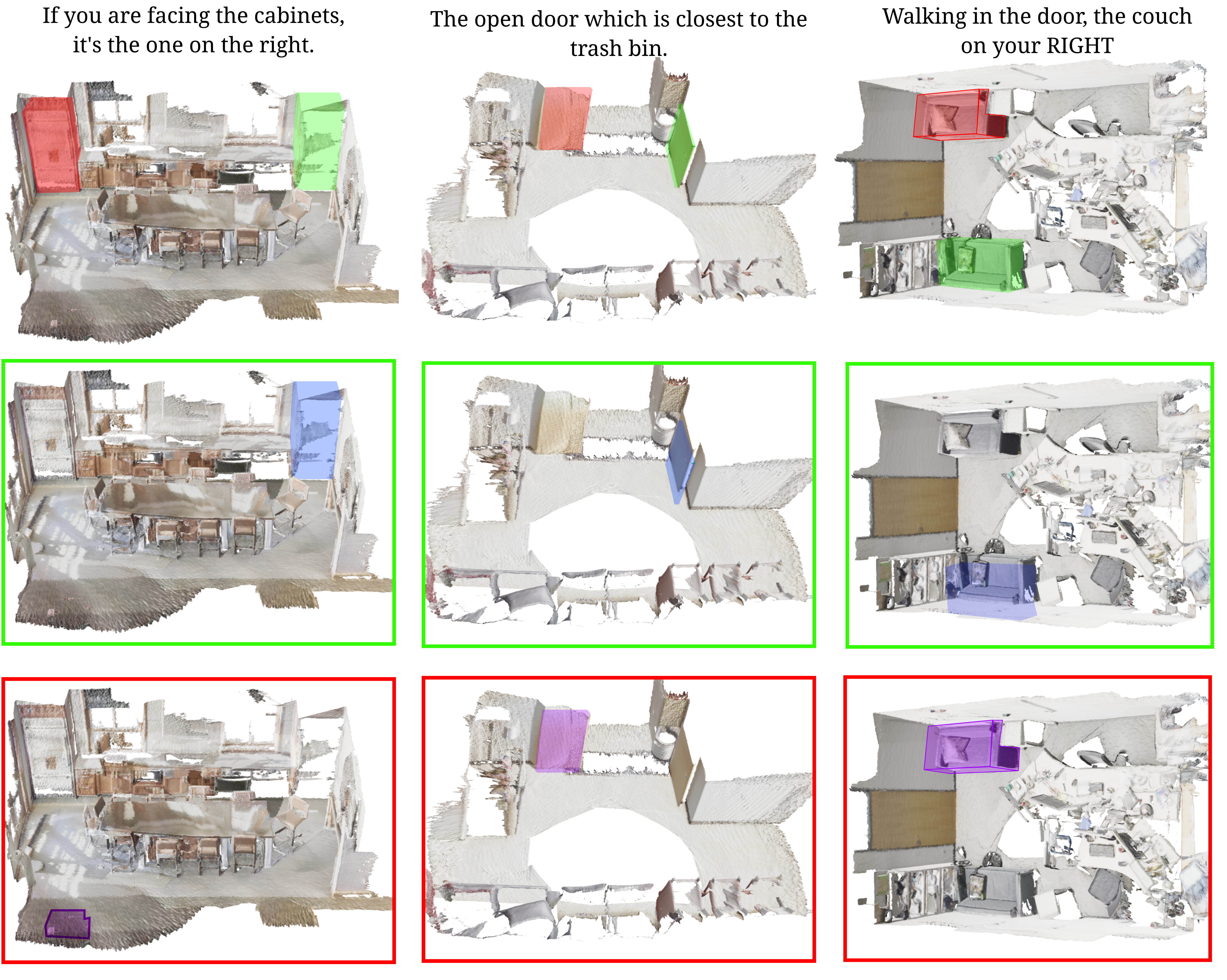}
\caption{Examples on the Nr3D dataset. The first row presents the ground-truth where the green boxes denote the target object and the red boxes denote distractor objects of the same class. The second row shows the predictions (blue boxes) from our proposed ViL3DRel model. The third row shows the predictions (purple boxes) from a baseline model without spatial self-attention and knowledge distillation.}
	\label{fig:supp_examples_cmpr}
\end{figure}

Figure~\ref{fig:supp_examples_cmpr} compares predictions by our proposed ViL3DRel model and the baseline model without the spatial self-attention and knowledge distillation.
We can see that our model is better at object perception (left example) while reasoning about different types of spatial relations such as relative distances (middle example) and orientations (right example).

\begin{figure}
    \includegraphics[width=\linewidth]{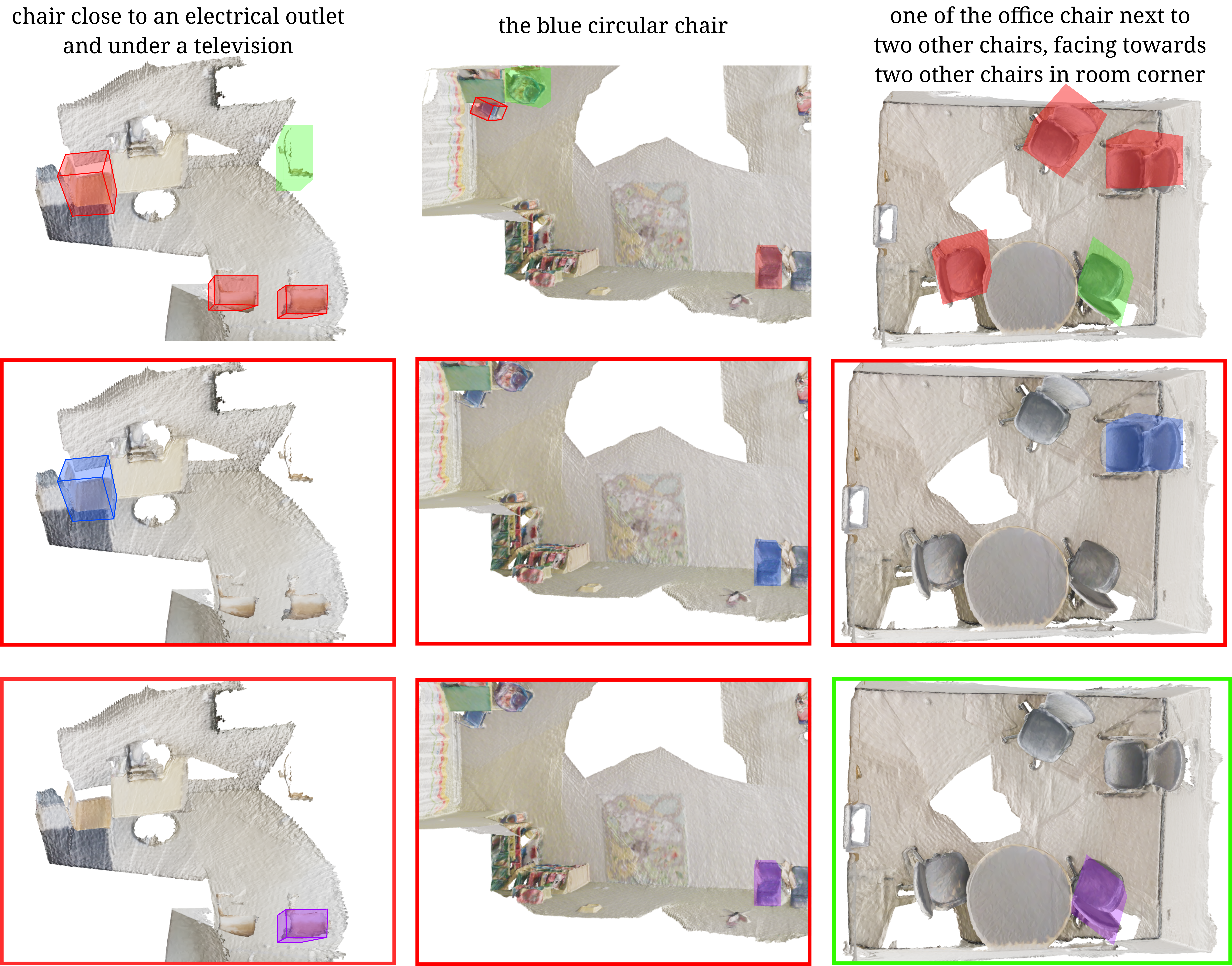}
	\caption{Failure cases of the proposed ViL3DRel model. The first row presents the ground-truth where the green boxes denote the target object and the red boxes denote distractor objects of the same class. The second row shows the predictions (blue boxes) from our proposed ViL3DRel model.}
	\label{fig:supp_failure_cases}
\end{figure}

In Figure~\ref{fig:supp_failure_cases}, we further provide some failure cases of the ViL3DRel model.
Example on the left is missing object proposals for the outlet and television. Since our model belongs to the two-stage method, its performance is highly dependent on the quality of object proposals in the first stage.
The middle example suggests that the current model still suffers from recognizing fine-grained attributes from the point clouds.
The right example further requires reasoning about object orientations, which might be hard to estimate from noisy point clouds.

\end{document}